\documentclass[conference]{IEEEtran}
\usepackage{times}

\usepackage[numbers]{natbib}
\usepackage{multicol}
\usepackage[bookmarks=true]{hyperref}

\usepackage{graphicx}
\usepackage{amsmath}
\usepackage{tabu}
\usepackage{multirow}
\usepackage{array}

\newcommand{\argmin}{\operatornamewithlimits{argmin}}
\begin{document}

\title{Tree Morphology for Phenotyping from\\Semantics-Based Mapping in Orchard Environments}

\author{Wenbo Dong and Volkan Isler\\
Department of Computer Science and Engineering, University of Minnesota, Twin Cities\\
Email: \{dongx358, isler\}@umn.edu}



%

\maketitle

\begin{abstract}
Measuring tree morphology for phenotyping is an essential but labor-intensive activity in horticulture. Researchers often rely on manual measurements which may not be accurate for example when measuring tree volume. Recent approaches on automating the measurement process rely on LIDAR measurements coupled with high-accuracy GPS. Usually each side of a row is reconstructed independently and then merged using GPS information. Such approaches have two disadvantages: (1)~they rely on specialized and expensive equipment, and (2)~since the reconstruction process does not simultaneously use information from both sides, side reconstructions may not be accurate. We also show that standard loop closure methods do not necessarily align tree trunks well. In this paper, we present a novel vision system that employs only an RGB-D camera to estimate morphological parameters. A semantics-based mapping algorithm merges the two-sides 3D models of tree rows, where integrated semantic information is obtained and refined by robust fitting algorithms. We focus on measuring tree height, canopy volume and trunk diameter from the optimized 3D model. Experiments conducted in real orchards quantitatively demonstrate the accuracy of our method.
\end{abstract}


\IEEEpeerreviewmaketitle

\section{Introduction} \label{sec:introduction}
The estimation of morphological parameters of fruit trees (such as tree height, canopy volume and trunk diameter) is important in horticultural science, and has become an important topic in precision agriculture~\cite{rosell2012review, tabbrobotic}. Accurate morphology estimation can help horticulturists study to what extent these parameters impact crop yield, health and development. For example, growers try different root stocks to figure out which one produces better yield per volume for a specific geographical area. They also measure parameters such as tree height or trunk diameter to model fruit production. This measurement process is labor-intensive and not necessarily accurate.

2D or 3D LIDAR scanning has proven to be a viable option for generating 3D models of trees~\cite{underwood2015lidar,bargoti2015pipeline}. Usually, LIDAR sensors are mounted on a vehicle moving along the alleys of the fruit orchard to vertically scan the side of the tree rows~\cite{mendez2014deciduous,underwood2016mapping}. To obtain the 3D point cloud by adding subsequent of 2D transects of laser scanning, the vehicle has to move with a steady velocity and along a linear track parallel to the tree row. However, these systems do not merge two scanned sides of trees. Morphological parameters are thus inaccurately computed by only scanning one side and multiplying by two or by adding the volumes of the two sides without merging them.
Generated two-sides point clouds can also be manually matched through CAD software~\cite{rosell2009obtaining}. However, tree models are partially misaligned from two sides due to accumulated errors of sensor poses during the movement. Even if position accuracy has been improved by combining Global Navigation Satellite System (GNSS) with LIDAR~\cite{del2015georeferenced}, the issue of accumulated orientation error still exists especially for large scale scanning. Furthermore, the combination of these two sensors (e.g. GR3 RTK GNSS and LMS500) is expensive and may not be affordable.

Cameras are low-cost, lightweight compared to LIDAR sensors. Vision-based 3D dense reconstruction, with the ability to provide quantitative information of every geometric detail of an object, is a promising alternative for accurate morphology measurement. Although time-of-flight~\cite{van2012spicy}, stereo-vision systems~\cite{bac2014stem} and depth sensors~\cite{wang2014size} have been used to estimate parameters of low-height plants, these approaches have been limited to indoor environments with controlled conditions, such as constant background and artificial illumination. We focus on the outdoor case in natural orchard environments.
\begin{figure}[t]
	\centering
	\includegraphics[width=0.99\columnwidth]{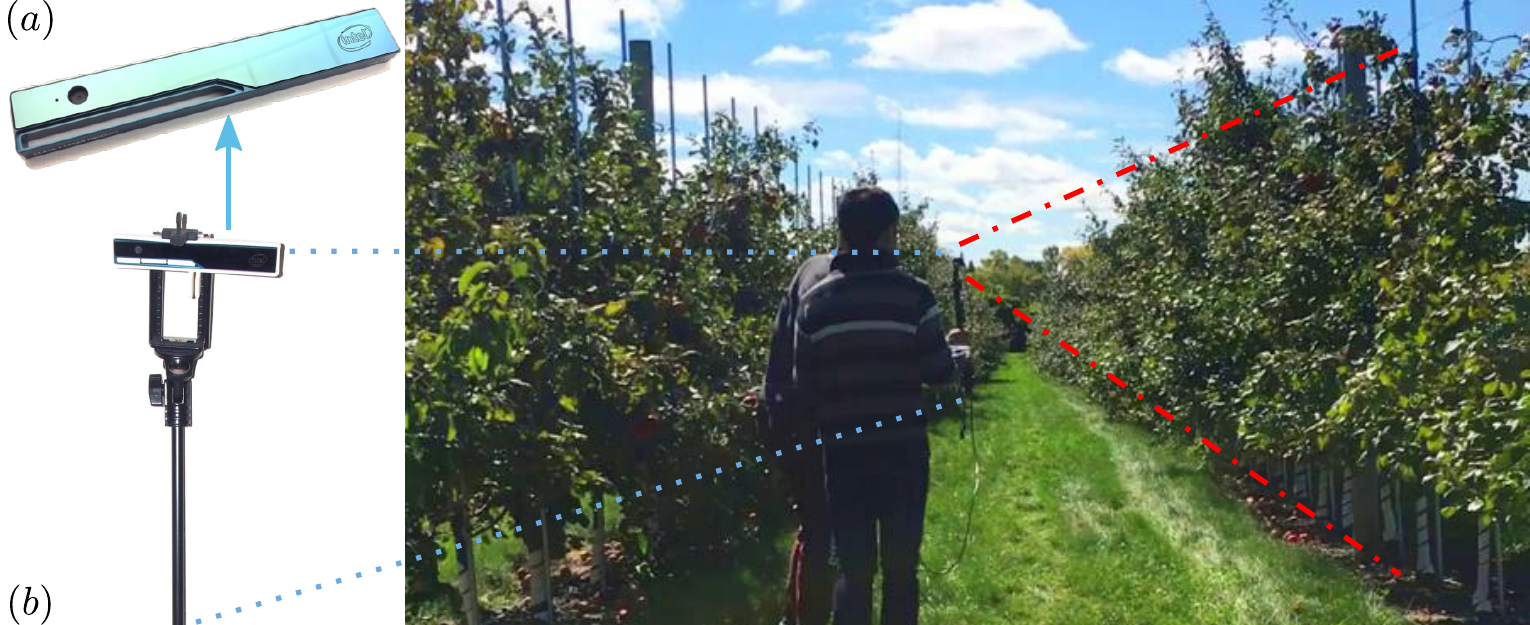}
	\caption{Overview of data capturing scenario. (a): The RGB-D camera (Intel RealSense R200). (b): The RGB-D sensor is mounted on a stick to capture data from either horizontal view or titled top-down view.}
	\label{fig:overview}
    \vspace*{-4mm}
\end{figure}

The goal of our work is to use RGB-D videos to reconstruct well-aligned 3D model of tree rows from images of both sides and estimate tree morphology. For a modern high-density orchard setting, it is not possible to perform mapping around each tree individually. Instead, two sides of tree rows are captured separately by a moving camera or in a loop trajectory. Obtaining accurate 3D models of fruit trees requires accurate camera poses, but estimating them reliably for long range  RGB-D videos is a difficult problem. Especially in orchard environments, good features cannot be stably tracked through long subsequent frames because of motion due to wind in the scene~\cite{dong2017linear}. Accumulated errors in camera poses will cause misalignment of tree models from both sides. As we show in Sec.~\ref{sec:technicalBackground}, state-of-the-art methods for volumetric fusion~\cite{newcombe2011kinectfusion}, Structure-from-Motion (SfM)~\cite{wu2013towards} and Simultaneous Localization and Mapping (SLAM)~\cite{mur2017orb} are not reliable enough for tree volume and trunk diameter estimation. Since there is nearly no overlap of canopy surface between two sides of tree rows, misalignment of tree models cannot be addressed by ICP-based methods~\cite{medeiros2017modeling} or semantic tracking in loop closure~\cite{bowman2017probabilistic}.

Our method relies on establishing semantic relationships between each of the two-sides and integrating tree morphology into the reconstruction system, which in turn outputs optimized morphological parameters.
Fig.~\ref{fig:overview} illustrates an overview of our data collection. To the best of our knowledge, it is the first vision system for accurate estimation of tree morphology in fruit orchards by using only an RGB-D camera. In summary, our work has the following key contributions:
\begin{itemize}
\item We present a novel mapping approach on RGB-D videos that can separately reconstruct 3D models of fruit trees from both sides and accurately merge them based on semantics, i.e. tree trunks and local ground patches.
\item We introduce robust fitting algorithms to estimate the initial trunk size and local planar ground for each tree.
\item We integrate tree-trunk diameters into semantic SfM to further localize trunks and local ground patches.
\item We measure tree height, tree volume and trunk diameter through automated segmentation for each tree based on optimized information of trunks and local grounds.
\end{itemize}

This paper is structured as follows. After discussing technical challenges, we introduce our proposed tree morphology estimation, followed by experimental results and a conclusion.


\vspace*{2mm}
\section{Technical Background} \label{sec:technicalBackground}
This section provides the problem formulation of tree morphology estimation with an overview of our system, and two main challenges of 3D reconstruction in orchard environments.

\subsection{Problem Formulation of Tree Morphology Estimation}

\begin{figure}[t]
	\centering
	\includegraphics[width=0.99\columnwidth]{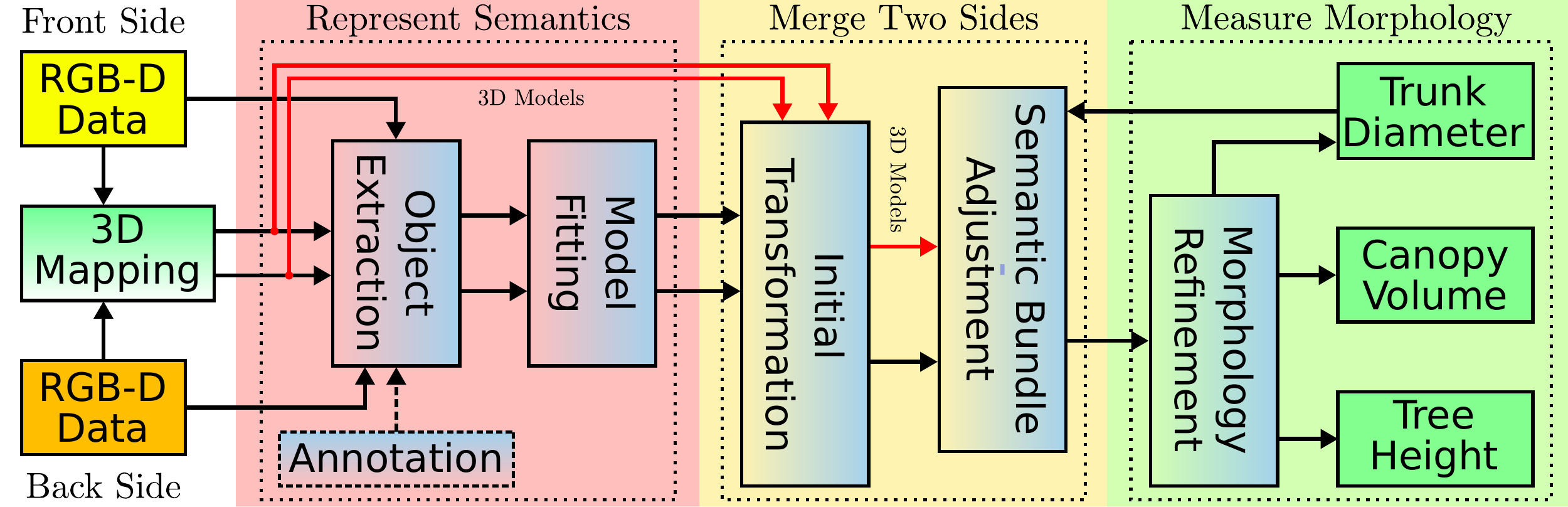}
	\caption{Overview of proposed system for tree morphology estimation. The trunk annotation (dashed line) in object extraction can be replace by trunk detection~\cite{bargoti2015pipeline} if without the need for trunk diameter estimation.}
	\label{fig:system}
	\vspace*{-5mm}
\end{figure}

Consider the problem of tree morphology estimation, in which a mobile camera separately moving along both sides of a tree row collects the RGB-D data of static landmarks (3D points and 3D objects, such as trunks and local grounds). 
The true models of the two sides  are related by a rigid transformation $\mathcal{T}$. Given a set of RGB-D measurements $\bar{\mathcal{X}}_k$ and object types $\mathcal{I}_j$, the task is to estimate the object poses $\mathcal{S}^{\mathcal{I}}_j$ with their sizes $\mathcal{D}^{\mathcal{I}}_j$, the transformation $\mathcal{T}$, along with the 3D point positions $\mathcal{X}_i$ and camera poses $\mathcal{C}_k$:
\begin{equation} \label{probForm}
\scalebox{0.97}
{$
\begin{aligned}
\argmin\limits_{\mathcal{S}^{\mathcal{I}}_j, \mathcal{D}^{\mathcal{I}}_j, \mathcal{T}, \mathcal{X}_i, \mathcal{C}_k} \sum\limits_{j} &\sum\limits_{k} \sum\limits_{i \in \mathcal{V}(j,k)} E_{\mathcal{S}} (\bar{\mathcal{X}}_k, \mathcal{T}, \mathcal{S}^{\mathcal{I}}_j, \mathcal{D}^{\mathcal{I}}_j, \mathcal{X}_i, \mathcal{C}_k) \\
+ &\sum\limits_{k} \sum\limits_{i \in \mathcal{V}(k)} E_{\mathcal{X}} (\bar{\mathcal{X}}_k, \mathcal{T}, \mathcal{X}_i, \mathcal{C}_k)
\end{aligned}
$} ,
\end{equation}
where $E_{\mathcal{S}}$ is the cost between a measured point and the object it belongs to, and $E_{\mathcal{X}}$ is the cost between a 3D point visible from a camera frame and its measurement. The proposed vision system for estimating tree morphology is illustrated in Fig.~\ref{fig:system}. 
The estimation procedure is divided into three steps explained in Sec.~\ref{sec:methodology}. We note that even though our approach  starts with two independent reconstructions of the two sides, it refines them based on semantic information.

\subsection{Technical Challenges} \label{subsec:technicalChallenges}
In modern orchards, fruit trees are highly packed in each row and connected by supporting wires (see Fig.~\ref{fig:overview}). Without enough separate space, it is not possible to individually perform surrounding RGB-D data collection around each tree. Instead, we collect side-view data of tree rows by moving the RGB-D camera along the path between the rows.
The rows can be hundreds of meters long. But the specific region of interest for a particular study can be only a subset of the row.
If we measure only this region from the two sides, the images across the sides may have no overlap. Alternatively, the entire row can be covered by following a loop around the row. In this section, we detail technical challenges associated with these two approaches.

\begin{figure}[t]
	\centering
	\includegraphics[width=0.99\columnwidth]{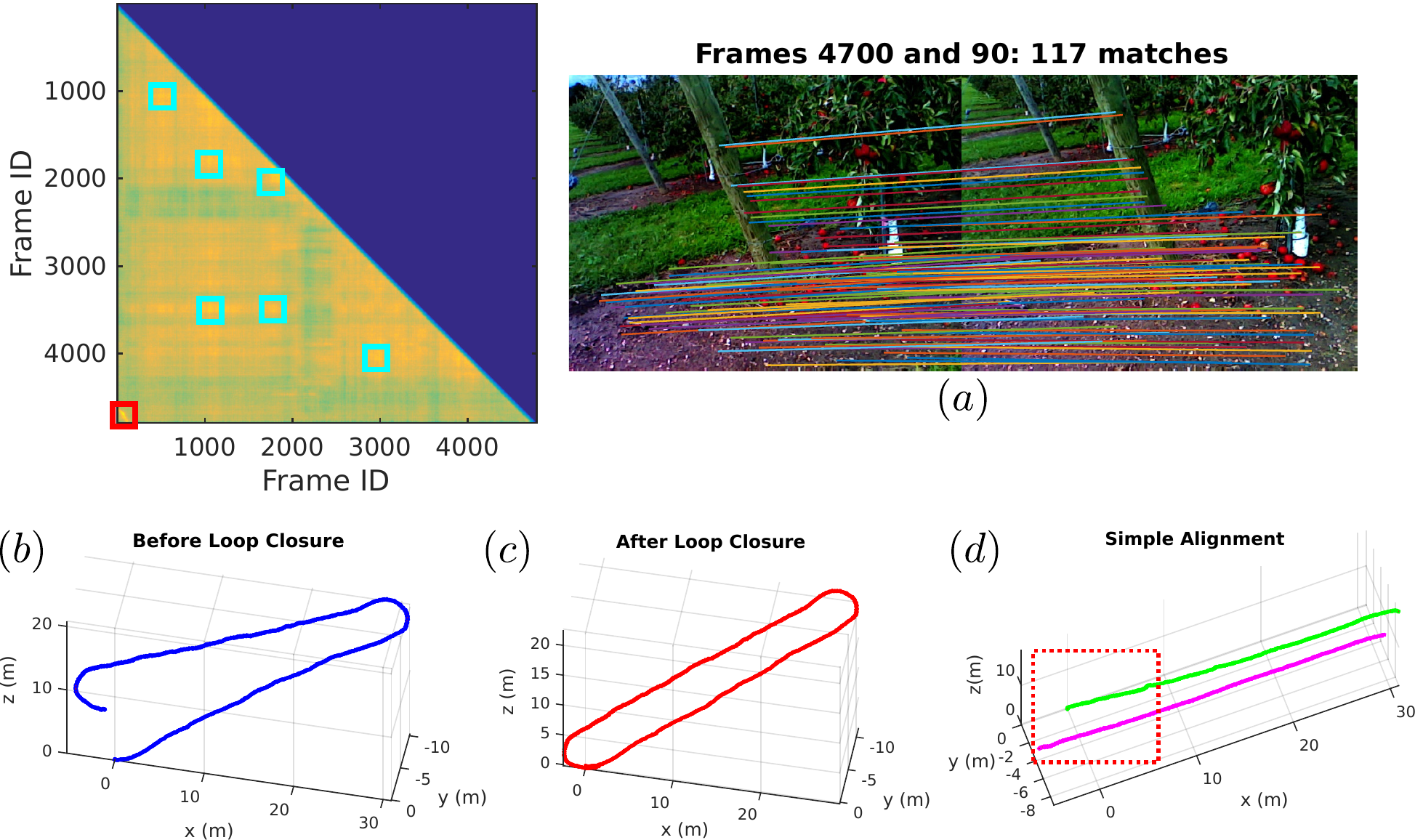}
	\caption{The score matrix between all image frames generated by using a BoW model. High similarities are marked by colored boxes. The correct loop detection is marked by the red box. (a): Feature matching between a pair of frames detected by loop closure. (b): Camera trajectory before loop closure. (c): Camera trajectory after loop closure. (d): Simple alignment of two-sides 3D models is not feasible: camera trajectories from both sides are diverged and marked by the red box.}
	\label{fig:loopClosure}
	\vspace*{-2mm}
\end{figure}

\begin{figure}[t]
	\centering
	\includegraphics[width=0.99\columnwidth]{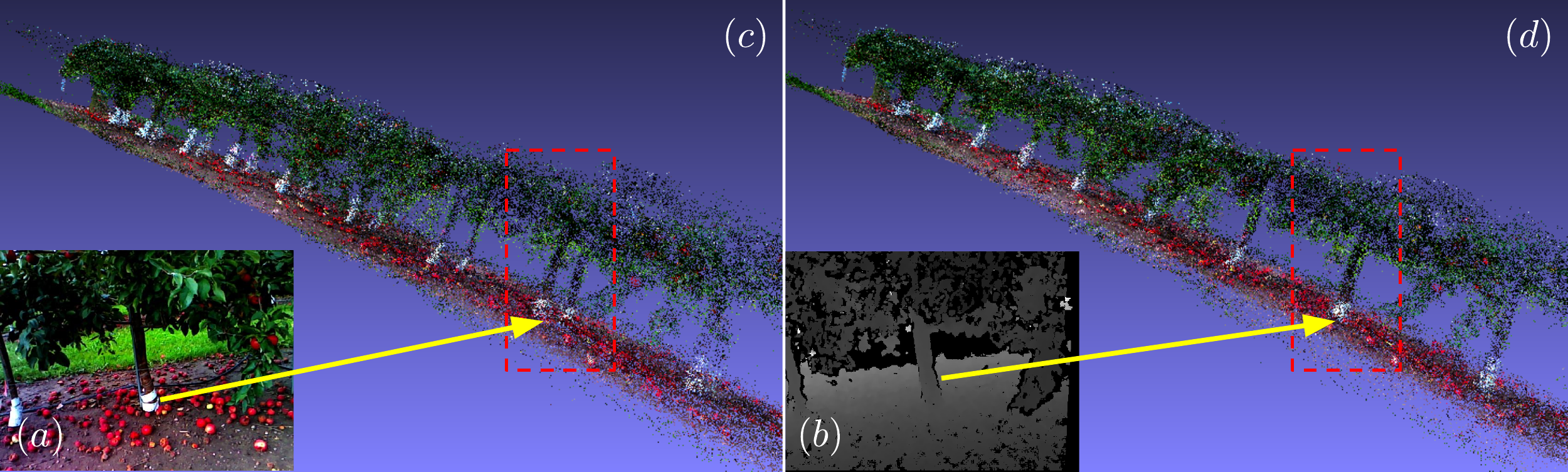}
	\caption{Even with loop closure, the 3D reconstruction of tree rows is not satisfactory: 3D models of tree trunks from both sides are misaligned. (a): The RGB image. (b): The depth image. (c): Misaligned trunks from both sides. (d): The 3D reconstruction is improved by integrating trunks information.}
	\label{fig:duplicated}
	\vspace*{-4mm}
\end{figure}

First, ORB-SLAM2~\cite{mur2017orb} is tested on our RGB-D data captured in a loop around a tree row to create the 3D model. Unlike indoor cases, image features in orchard environments are unstable due to wind effect and thus hard to track across multiple frames, which causes the SLAM algorithm frequently getting lost. On the other hand, loop detection is not reliable because of high similarity between fruit trees of the same type (see Fig.~\ref{fig:loopClosure}). With correct loop closure, the 3D dense reconstruction of the tree row from both sides is generated by converting depth maps into point clouds based on the optimized camera trajectory from the SLAM output. From Fig.~\ref{fig:duplicated}, we observe that although the loop is correctly closed the 3D model of the tree row is not satisfactory. The 3D dense reconstruction has separated trunks since there is no data overlap between both sides of the tree row. Measuring tree morphology based on inaccurate models is problematic, especially canopy volume and trunk diameter estimation.

For the data separately captured from both sides, simple alignment of two-sides 3D models can be performed by estimating the rigid transformation based on the trunks information. However, due to accumulated errors of camera poses, some trees are well-aligned from both sides (with parallel camera trajectories) while the rest are misaligned (see Fig.~\ref{fig:loopClosure}d). Fig.~\ref{fig:duplicated}d implies that two-sides 3D reconstruction should be further optimized based on semantic information to correct camera trajectories. Standard SfM algorithm~\cite{lowe2004distinctive} often fails to close loops when dealing with view-invariant feature matching, and may converge to a local minimum. Hence, we adjust the single-side 3D reconstruction by integrating essential elements from SLAM and SfM algorithms.

\subsection{Single-Side Reconstruction} \label{subsec:singleSideBA}
In this section, we present the proposed approach for initially reconstructing each side independently using established techniques.
For each pair of consecutive frames, the relative rigid transformation is calculated by applying a RANSAC-based three-point-algorithm~\cite{forsyth2011computer} on the SIFT matches~\cite{lowe2004distinctive} with valid depth values. Pairwise Bundle Adjustment (BA) is performed to optimize the relative transformation and 3D locations of matches by minimizing 2D reprojection errors. For loop detection, we build a Bag of Words (BoW) model~\cite{sivic2009efficient} to characterize each frame with a feature vector, which is calculated based on different frequencies of visual words. The score matrix is obtained by computing the dot products between all pairs of feature vectors (see Fig.~\ref{fig:loopClosure}). Possible loop pairs are first selected by a high score threshold and then tested by RANSAC-based pose estimation whether a reasonable number of good matches are obtained (e.g. 100 SIFT matches). Loop pairs are thus accurately detected and linked with pairs of consecutive frames by covisibility graph. Loop detection allows us capture each single tree back and forth from different views on a single side.

For each frame in consecutive pairs, we first perform local BA to optimize its local frames which have common features. To effectively close the loop, pose graph optimization~\cite{strasdat2010scale} is then performed followed by global BA to finally optimize all camera poses and 3D points. Given the fact that depth maps in outdoor cases are generated by infrared stereo cameras, we integrate 3D errors information into the objective function of bundle adjustment as follows:
\begin{equation} \label{objectBA}
\begin{gathered}
\argmin_{\mathbf{R}_c, \mathbf{t}_c, \mathbf{X}_p} J = \sum_{c} \sum_{p \in \mathcal{V}(c)} \rho\left( E_o(c,p) \right) + \rho\left( E_i(c,p) \right) \\
E_o(c,p) = \| ^c\bar{\mathbf{x}}_p - \mathbf{K}_o [\mathbf{R}_c | \mathbf{t}_c ] \mathbf{X}_p \|^2 \\
E_i(c,p) = \| \mathbf{K}_i [\mathbf{R}_i | \mathbf{t}_i ] ^c\bar{\mathbf{X}}_p - \mathbf{K}_i [\mathbf{R}_i | \mathbf{t}_i ] [\mathbf{R}_c | \mathbf{t}_c ] \mathbf{X}_p \|^2
\end{gathered} ,
\end{equation}
where $\rho$ is the robust Huber cost function~\cite{huber1992robust}, $\mathbf{K}_o$ and $\mathbf{K}_i$ are intrinsics matrices of the RGB camera and the left infared camera, $[\mathbf{R}_i | \mathbf{t}_i ]$ is the relative transformation between these two cameras, $[\mathbf{R}_c | \mathbf{t}_c ]$ is the RGB camera pose, $\mathbf{X}_p$ is the 3D location of a point visible from the camera frame $c$, and $^c\bar{\mathbf{x}}_p$ and $^c\bar{\mathbf{X}}_p$ are the observed 2D feature and 3D location in the RGB camera frame, respectively.


%

\section{Methodology} \label{sec:methodology}
In this section, we present our main technical contribution: merging and refining the reconstructions of the two sides using semantic information.
The proposed method consists of three steps (see Fig.~\ref{fig:system}).

\subsection{Trunk Fitting and Local Ground Estimation} \label{subsec:trunkGround}
Accurate geometry estimation relies on good depth maps. The raw depth maps are usually noisy, especially in orchard environments. The big uncertainty of depth values around frequent occlusions between trees and leaves causes generated 3D points floating in the air~\cite{sotoodeh2006outlier}.
We first improve the depth map using the Truncated Signed Distance Function (TSDF)~\cite{curless1996volumetric}  to accumulate depth values from nearby frames (e.g. 10 closest frames) with the camera poses obtained in Sec.~\ref{subsec:singleSideBA}. The pixel value of the raw depth is ignored if it is largely different from the corresponding value in the fused depth obtained by ray casting. A floating pixel removal filter~\cite{sotoodeh2006outlier} is further applied to eliminate any pixel of the raw depth that has no nearby 3D points within a certain distance threshold.
\begin{figure}[t]
	\centering
	\includegraphics[width=0.99\columnwidth]{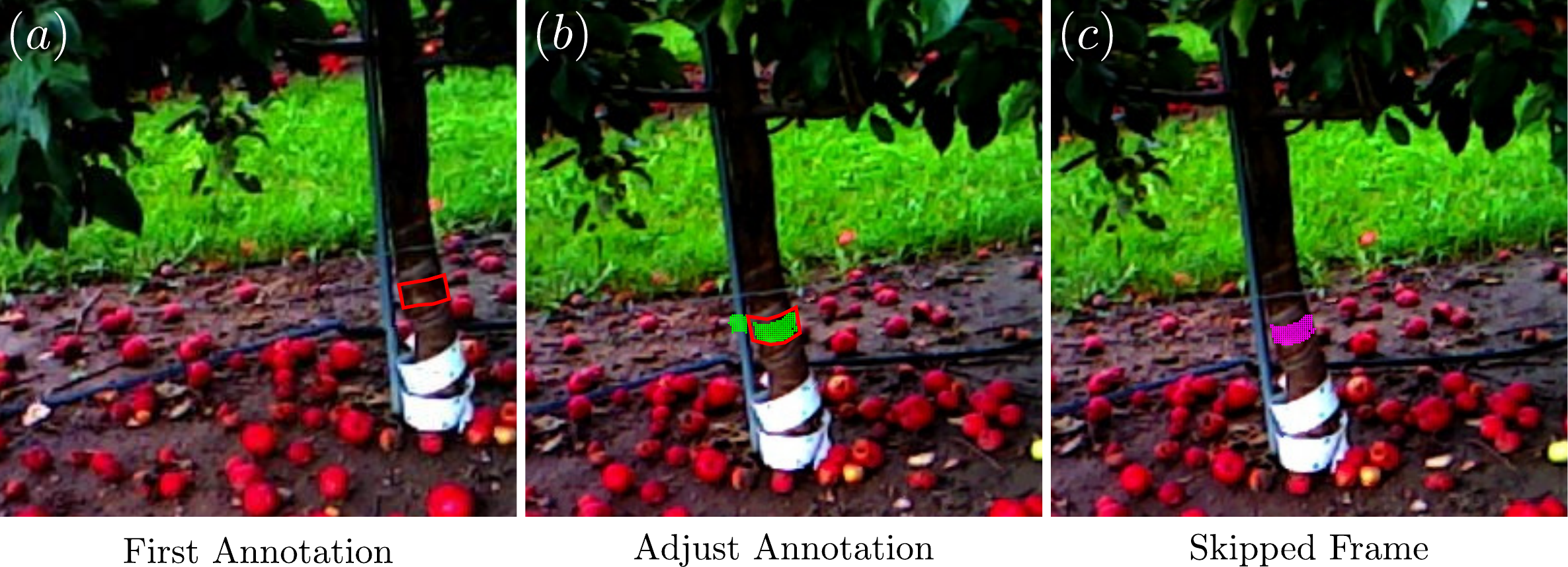}
	\caption{Trunk annotation. (a): The trunk is first annotated by a red polygon in frame 694. (b): The red polygon is adjusted in frame 718 if the depth pixels (green) selected by the projected region are not satisfactory. (c): The frame 719 is skipped without annotation if the depth pixels (magenta) are within the trunk region-of-interest.}
	\label{fig:annotation}
	\vspace*{-4mm}
\end{figure}

\subsubsection{Trunk Region-of-Interest Selection}
Horticulturists typically measure the trunk diameter of a fruit tree at the height about a fist width above the graft union. Without a consistent rule, we create a annotation tool for horticulturists to mark the trunk region-of-interest.

As shown in Fig.~\ref{fig:annotation}, a user first needs to annotate the region for measuring the trunk by a polygon in the frame $c$ around the best view of the tree. 3D points of this frame generated based on the polygon mark of the depth image are then projected to the next frame $c+1$ and enclosed by a convex polygon. Depth pixels of frame $c+1$ are then highlighted within this convex polygon to allow the user checking whether the highlighted region is still correct. The user needs to create a new annotation if the projected region is not satisfactory due to errors of camera poses or depth values. The new annotated polygon is updated to create projected regions for the following frames. The nearby frames usually have correct projected regions and are thus skipped without any annotation.

\subsubsection{Trunk Cylinder} \label{subsubsec:trunk}
For annotated frames, a 3D point cloud of the trunk in frame $c$ is generated and filtered by taking the intersection of polygon masks with two nearby frames $c-1$ and $c+1$. We aim to fit the 3D points to a cylinder $d$ parameterized by its axis $^c\mathbf{n}_d$, center $^c\mathbf{O}_d$ and radius $^cr_d$. The height $^ch_d$ of the cylinder is determined by the bounding box of 3D points along $^c\mathbf{n}_d$. 
\begin{figure}[t]
	\centering
	\includegraphics[width=0.99\columnwidth]{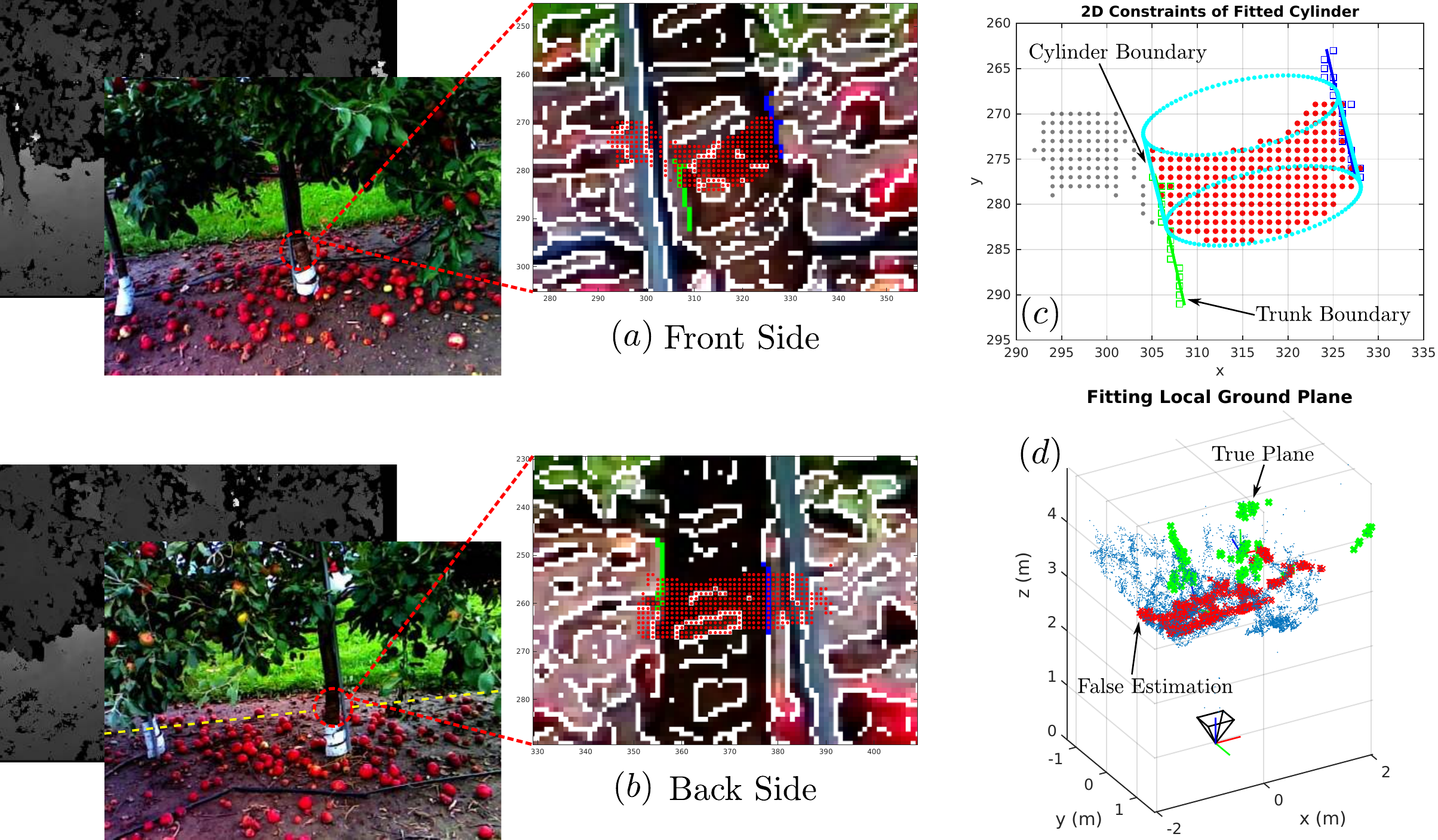}
	\caption{Trunk fitting and local ground estimation from both sides. The estimated plane from the front side helps the user locate the height of the trunk (yellow line) from the back side. (a) and (b): Trunk boundaries are detected (green and blue) from Canny edges. The depth pixels (red) are selected by projected convex polygon. (c): The 2D constraints of cylinder fitting from the front side with marked inliers (red) and outliders (gray) from the depth. (d): Without trunk information, standard plane estimation outputs a wrong plane (red), while the true ground (green) is estimated using proposed algorithm.}
	\label{fig:fittingEstimation}
	\vspace*{-4mm}
\end{figure}

A good cylinder model should not only fit the most of 3D points but also obtain a reasonable size from the image. To robustly model the cylinder, we integrate 2D constraints into a RANSAC scheme~\cite{fischler1981random} with the nine-point algorithm~\cite{beder2006direct}. Specifically, Canny edge detection~\cite{canny1986computational} is first performed (see Fig.~\ref{fig:fittingEstimation}). Based on the silhouette of the annotated polygon, two trunk boundaries are detected and fitted to lines $\mathbf{l}_a$ and $\mathbf{l}_b$ using the total least squares method~\cite{golub1980analysis}. Two cylinder boundaries $\mathbf{l}_{\alpha}$ and $\mathbf{l}_{\beta}$ are extracted by projecting the circles of two cylinder ends onto the image. The trunk cylinder in frame $c$ is further optimized by minimizing the cost function
\begin{equation} \label{objectCylinder}
\scalebox{0.98}
{$
\argmin\limits_{^c\mathbf{n}_d, ^c\mathbf{O}_d, ^cr_d} \sum\limits_{p} e_d^2(^c\mathbf{X}_p, d) + \lambda \left( \| \hat{\mathbf{l}}_{\alpha} - \hat{\mathbf{l}}_a \|^2 + \| \hat{\mathbf{l}}_{\beta} - \hat{\mathbf{l}}_b \|^2 \right)
$} ,
\end{equation}
where $e_d$ is the distance function of a 3D point $^c\mathbf{X}_p$ to the cylinder, and $\hat{\mathbf{l}}_{\alpha}$, $\hat{\mathbf{l}}_{\beta}$, $\hat{\mathbf{l}}_a$, and $\hat{\mathbf{l}}_b$ are normalized unit vectors. The trunk in frame $c$ is thus described by the cylinder axis $^c\mathbf{n}_d$ and the origin $^c\mathbf{O}_d$.

\subsubsection{Local Ground Plane} \label{subsubsec:plane}
Without loss of generality, the local ground of a tree is assumed as a plane defined by its normal $^c\mathbf{n}_p$ and center $^c\mathbf{O}_p$ in frame $c$. Unlike trunk annotation, only frame number is recorded for plane estimation. However, it is not always the case that the majority of 3D points are from the ground, which highly depends on the scene and the camera view. The standard RANSAC-based method fails to detect the ground plane (see Fig.~\ref{fig:fittingEstimation}d). We modify the degenerate condition of the RANSAC by using the prior information of the trunk axis $^c\mathbf{n}_d$ transformed from the closest annotated frame: $^c\mathbf{n}_p$ should roughly align with $^c\mathbf{n}_d$, and the estimated plane should be on the boundary of all 3D points along $^c\mathbf{n}_p$ within the distance threshold $t_s$. The local ground in frame $c$ is thus defined by the plane normal $^c\mathbf{n}_p$ and the origin $^c\mathbf{O}_p$. Local ground estimation from the front side can further help annotations for the back side (see Fig.~\ref{fig:fittingEstimation}).

\begin{figure}[t]
	\centering
	\includegraphics[width=0.99\columnwidth]{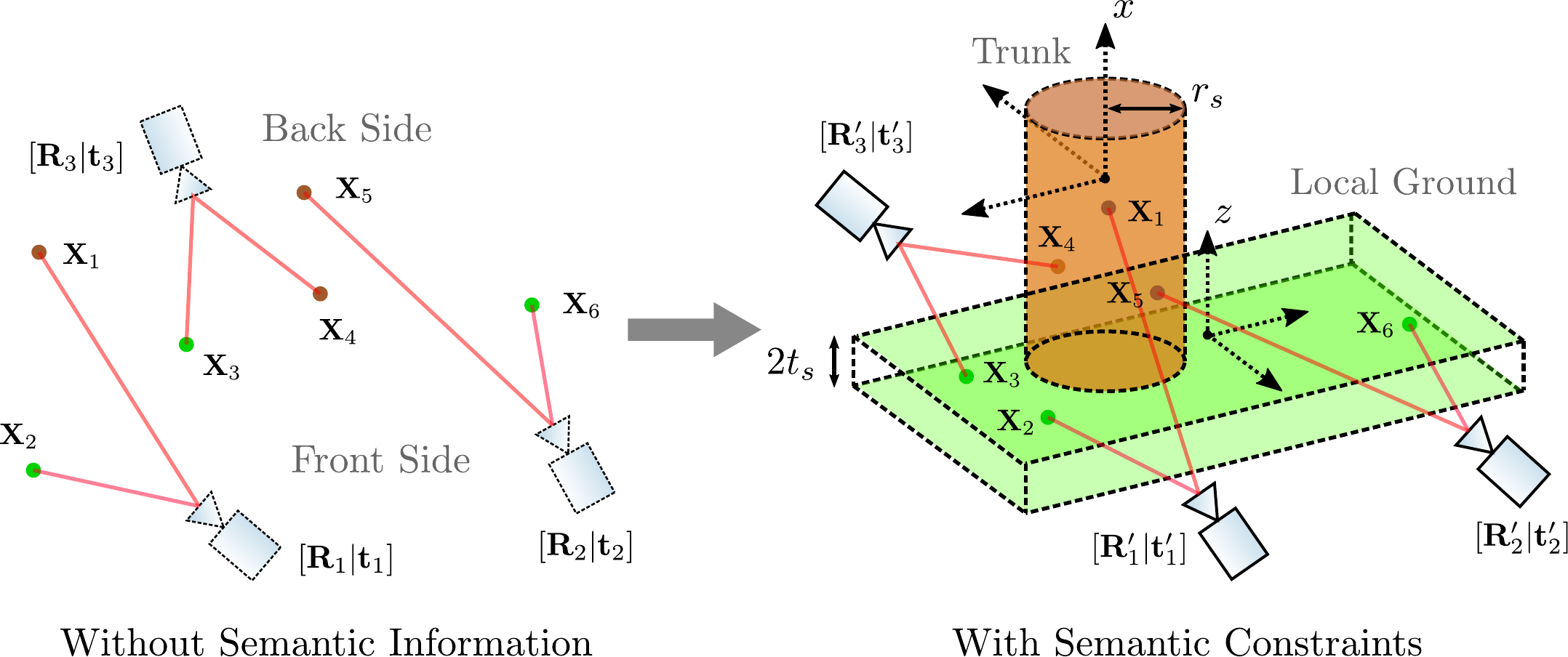}
	\caption{The scheme of semantic bundle adjustment. With semantic constraints, 3D points belonging to the same object are adjusted to fit onto the shape together with the camera poses corrected simultaneously.}
	\label{fig:semanticBA}
	\vspace*{-4mm}
\end{figure}

\subsection{Merging Two-Sides 3D Reconstruction}
For a tree row, the front-side and back-side reconstructions are expressed in their own frames ${\mathcal{F}}$ and ${\mathcal{B}}$, respectively. The goal is to first align two-sides reconstructions by estimating the initial transformation $[^\mathcal{F}_\mathcal{B}\mathbf{R} | ^\mathcal{F}_\mathcal{B}\mathbf{t}]$, and further optimize the 3D reconstruction based on semantic information.

\subsubsection{Initial Transformation}
From a geometric view, to align the 3D models of a tree row from both sides, at least two annotated trunks and one estimated local ground are required. 3D models are first constrained on the local ground plane. The translation and rotation along the ground plane are further constrained by two trunk-cylinders. Multiple trunks and local grounds can provide us a robust solution. In Sec.~\ref{subsec:trunkGround}, an $i$-th annotated trunk from two-sides annotated views is described by its cylinder axes $^{\mathcal{F}}\mathbf{n}_d^i$ and $^{\mathcal{B}}\mathbf{n}_d^i$ with a unit length, and its origins $^{\mathcal{F}}\mathbf{O}_d^i$ and $^{\mathcal{B}}\mathbf{O}_d^i$. Similarly, a $j$-th estimated local ground is described by its plane normals $^{\mathcal{F}}\mathbf{n}_p^j$ and $^{\mathcal{B}}\mathbf{n}_p^j$, and its origins $^{\mathcal{F}}\mathbf{O}_p^j$ and $^{\mathcal{B}}\mathbf{O}_p^j$.

First, cylinder axes and plane normals in ${\mathcal{B}}$ after the relative transformation must be equal to their corresponding ones in ${\mathcal{F}}$. Then, the first two constraints have the form
\begin{equation} \label{constraint1}
\begin{cases}
^\mathcal{F}_\mathcal{B}\mathbf{R} \cdot ^{\mathcal{B}}\mathbf{n}_d^i = ^{\mathcal{F}}\mathbf{n}_d^i \\
^\mathcal{F}_\mathcal{B}\mathbf{R} \cdot ^{\mathcal{B}}\mathbf{n}_p^j = ^{\mathcal{F}}\mathbf{n}_p^j
\end{cases} .
\end{equation}
Second, the origins of cylinders in ${\mathcal{B}}$ transformed to ${\mathcal{F}}$ should lie on the same axis-line. Then, the cross product between the cylinder axis and the difference of two-sides origins should be a zero vector
\begin{equation} \label{constraint2}
^{\mathcal{F}}\mathbf{n}_d^i \times \left( ^\mathcal{F}_\mathcal{B}\mathbf{R} \cdot ^{\mathcal{B}}\mathbf{O}_d^i + ^\mathcal{F}_\mathcal{B}\mathbf{t} - ^{\mathcal{F}}\mathbf{O}_d^i \right) = \mathbf{0} .
\end{equation}
At last, the origins of local planes in ${\mathcal{B}}$ after the transformation to ${\mathcal{F}}$ must lie on  the same plane. Thus, the dot product between the plane normal and the difference of two-sides origins should be zero
\begin{equation} \label{constraint3}
^{\mathcal{F}}\mathbf{n}_p^j \cdot \left( ^\mathcal{F}_\mathcal{B}\mathbf{R} \cdot ^{\mathcal{B}}\mathbf{O}_p^j + ^\mathcal{F}_\mathcal{B}\mathbf{t} - ^{\mathcal{F}}\mathbf{O}_p^j \right) = 0 .
\end{equation}

Following the order of constraints above, Eqs.~(\ref{constraint1})-(\ref{constraint3}) can be rearranged into a system of $\mathbf{A} \mathbf{x} = \mathbf{b}$ by treating each element of $[^\mathcal{F}_\mathcal{B}\mathbf{R} | ^\mathcal{F}_\mathcal{B}\mathbf{t}]$ as unknowns, where $^{\mathcal{B}}\mathbf{n}_d^i = [n^d_1, n^d_2, n^d_3]^{\top}$, $^{\mathcal{F}}\mathbf{n}_d^i = [{n^{\prime}}^d_1, {n^{\prime}}^d_2, {n^{\prime}}^d_3]^{\top}$, $^{\mathcal{B}}\mathbf{n}_p^j = [n^p_1, n^p_2, n^p_3]^{\top}$, and $^{\mathcal{F}}\mathbf{n}_p^j = [{n^{\prime}}^p_1, {n^{\prime}}^p_2, {n^{\prime}}^p_3]^{\top}$ for the axes, and the elements of origins have the similar form. Here, the matrix $\mathbf{A}$ and vector $\mathbf{b}$ are
\begin{equation}
\begin{gathered}
\scalebox{0.47}
{$
\begin{bmatrix}
n^d_1 & 0 & 0 & n^d_2 & 0 & 0 & n^d_3 & 0 & 0 & 0 & 0 & 0 \\
0 & n^d_1 & 0 & 0 & n^d_2 & 0 & 0 & n^d_3 & 0 & 0 & 0 & 0 \\
0 & 0 & n^d_1 & 0 & 0 & n^d_2 & 0 & 0 & n^d_3 & 0 & 0 & 0 \\
n^p_1 & 0 & 0 & n^p_2 & 0 & 0 & n^p_3 & 0 & 0 & 0 & 0 & 0 \\
0 & n^p_1 & 0 & 0 & n^p_2 & 0 & 0 & n^p_3 & 0 & 0 & 0 & 0 \\
0 & 0 & n^p_1 & 0 & 0 & n^p_2 & 0 & 0 & n^p_3 & 0 & 0 & 0 \\
0 & -{n^{\prime}}^d_3 o^d_1 & {n^{\prime}}^d_2 o^d_1 & 0 & -{n^{\prime}}^d_3 o^d_2 & {n^{\prime}}^d_2 o^d_2 & 0 & -{n^{\prime}}^d_3 o^d_3 & {n^{\prime}}^d_2 o^d_3 & 0 & -{n^{\prime}}^d_3 & {n^{\prime}}^d_2 \\
{n^{\prime}}^d_3 o^d_1 & 0 & -{n^{\prime}}^d_1 o^d_1 & {n^{\prime}}^d_3 o^d_2 & 0 & -{n^{\prime}}^d_1 o^d_2 & {n^{\prime}}^d_3 o^d_3 & 0 & -{n^{\prime}}^d_1 o^d_3 & {n^{\prime}}^d_3 & 0 & -{n^{\prime}}^d_1 \\
-{n^{\prime}}^d_2 o^d_1 & {n^{\prime}}^d_1 o^d_1 & 0 & -{n^{\prime}}^d_2 o^d_2 & {n^{\prime}}^d_1 o^d_2 & 0 & -{n^{\prime}}^d_2 o^d_3 & {n^{\prime}}^d_1 o^d_3 & 0 & -{n^{\prime}}^d_2 & {n^{\prime}}^d_1 & 0 \\
{n^{\prime}}^p_1 o^p_1 & {n^{\prime}}^p_2 o^p_1 & {n^{\prime}}^p_3 o^p_1 & {n^{\prime}}^p_1 o^p_2 & {n^{\prime}}^p_2 o^p_2 & {n^{\prime}}^p_3 o^p_2 & {n^{\prime}}^p_1 o^p_3 & {n^{\prime}}^p_2 o^p_3 & {n^{\prime}}^p_3 o^p_3 & {n^{\prime}}^p_1 & {n^{\prime}}^p_2 & {n^{\prime}}^p_3
\end{bmatrix}
$} \\ 
\scalebox{0.47}
{$
\begin{bmatrix}
{n^{\prime}}^d_1 & {n^{\prime}}^d_2 & {n^{\prime}}^d_3 & {n^{\prime}}^p_1 & {n^{\prime}}^p_2 & {n^{\prime}}^p_3 & {n^{\prime}}^d_2{o^{\prime}}^d_3 - {n^{\prime}}^d_3{o^{\prime}}^d_2 & {n^{\prime}}^d_3{o^{\prime}}^d_1 - {n^{\prime}}^d_1{o^{\prime}}^d_3 & {n^{\prime}}^d_1{o^{\prime}}^d_2 - {n^{\prime}}^d_2{o^{\prime}}^d_1 & {n^{\prime}}^p_1{o^{\prime}}^p_1 + {n^{\prime}}^p_2{o^{\prime}}^p_2 + {n^{\prime}}^p_3{o^{\prime}}^p_3
\end{bmatrix}^{\top}
$}
\end{gathered} ,
\end{equation}
respectively, and $\mathbf{x} = [\mathbf{r}_1^{\top}, \mathbf{r}_2^{\top}, \mathbf{r}_3^{\top}, ^\mathcal{F}_\mathcal{B}\mathbf{t}^{\top}]^{\top}$ with $\mathbf{r}_1$, $\mathbf{r}_2$ and $\mathbf{r}_3$ as three columns of $^\mathcal{F}_\mathcal{B}\mathbf{R}$.
\begin{figure}[t]
	\centering
	\includegraphics[width=0.99\columnwidth]{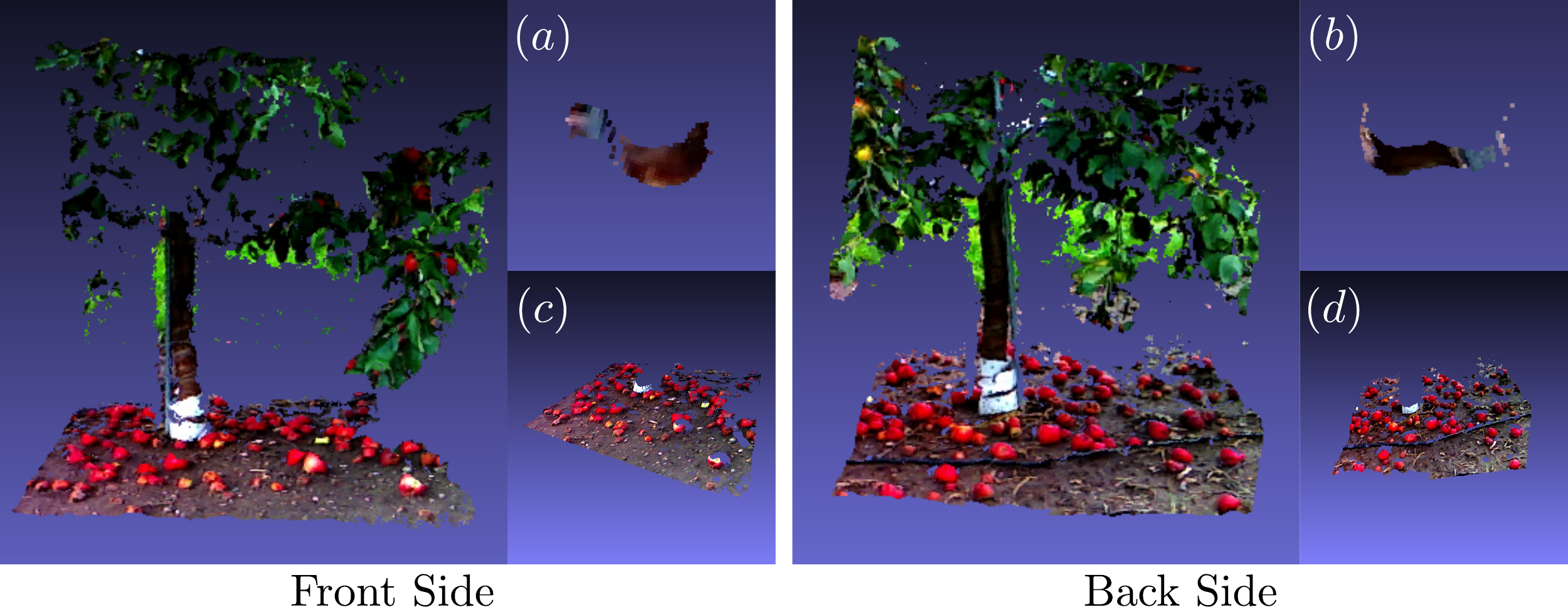}
	\caption{Front-side and back-side volumetric fusion using nearby frames. (a) and (b): Extracted 3D models of the trunk from both sides. (c) and (d): Extracted 3D models of the local ground from both sides.}
	\label{fig:fittingOverview}
	\vspace*{-4mm}
\end{figure}

We solve the system with multiple cylinders and planes for the least squares solution. The solution of $^\mathcal{F}_\mathcal{B}\mathbf{R}$ may not meet the properties of an orthonormal matrix, but can be computed to approximate a rotation matrix by minimizing the Frobenius norm of their difference~\cite{golub2012matrix}. An accurate initial value can be obtained from an analytical solution by using the resultant of polynomials~\cite{dong2016novel}. With multiple pairs of cylinders and planes from both sides, we formulate an optimization problem
\begin{equation}
\scalebox{0.93}
{$
\argmin\limits_{^\mathcal{F}_\mathcal{B}\mathbf{R}, ^\mathcal{F}_\mathcal{B}\mathbf{t}} \sum\limits_{i} \left( \|\mathbf{e}_1(i) \|^2 + |\ \mathbf{e}_3(i) \|^2 \right) + \sum\limits_{j} \left( \| \mathbf{e}_2(j) \|^2 + e_4^2(j) \right)
$} ,
\end{equation}
where $\mathbf{e}_1$, $\mathbf{e}_2$, $\mathbf{e}_3$ and $e_4$ are residuals of Eqs.~(\ref{constraint1})-(\ref{constraint3}). The solution is further refined using the Levenberg-Marquard (LM) method~\cite{levenberg1944method,marquardt1963algorithm} with the rotation represented by the Rodrigues' formula~\cite{rodrigues1840lois}.

\subsubsection{Semantic Bundle Adjustment} \label{subsubsec:SBA}
To address the issue of accumulated errors of camera poses in Fig.~\ref{fig:loopClosure}d, two-sides 3D reconstructions after initial alignment need to be further optimized. Intuitively, semantic information, i.e. trunks and local grounds, integrated in bundle adjustment will tune camera poses and 3D feature points until reasonable semantic conditions are reached. Specifically, two halves of a trunk from both sides should be well-aligned, and two-sides local grounds of a tree should refer to the same one (see Fig.~\ref{fig:semanticBA}).

\begin{figure}[t]
	\centering
	\includegraphics[width=0.99\columnwidth]{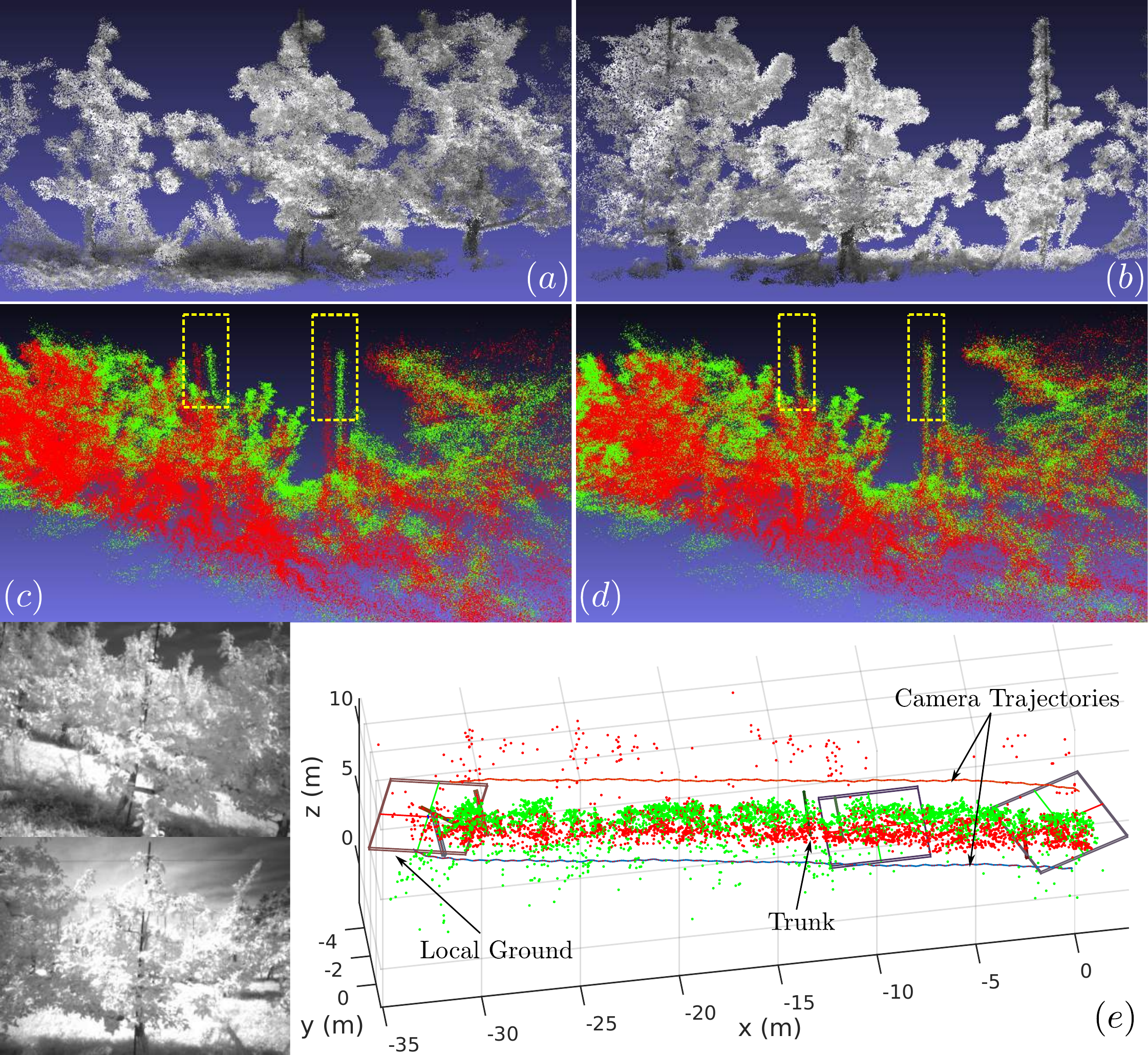}
	\caption{Merging 3D reconstruction of fruit trees for canopy volume estimation. (a) and (b): The 3D model viewed from both sides. (c): Some trunks are still misaligned after initial transformation. (d): Misalignments are eliminated by semantic BA. (e): 3D features from both sides are shown with camera poses  and semantic information (captured by stereo infrared cameras).}
	\label{fig:showSBA}
	\vspace*{-4mm}
\end{figure}

Technically, a semantic object with index $s$ is characterized by its unique pose $[\mathbf{R}_s | \mathbf{t}_s ]$ in the world frame and its 3D shape $\mathbf{b}_s$. For a cylinder object, the shape is represented by its $x$-axis (as the cylinder axis), origin and a radius $r_s$. For a plane object, the shape is described by its $z$-axis (as the plane normal), origin and a threshold $t_s$ for bounding an interval along the plane normal. The cylinder radius $r_s$ and the plane-interval threshold $t_s$ are automatically determined by the fitting algorithms in Sec.~\ref{subsubsec:diameter} and Sec.~\ref{subsubsec:plane}, respectively. As a 3D feature point, the orientation $\mathbf{R}_s$ and the position $\mathbf{t}_s$ of an object are unknown and to be estimated by semantic bundle adjustment.

Given the correspondences of objects between two sides, the objective function of semantic bundle adjustment is as follows
\vspace*{-3mm}
\begin{equation} \label{objectSBA}
\scalebox{0.97}
{$
\begin{gathered}
\argmin_{\mathbf{R}_c, \mathbf{t}_c, \mathbf{R}_s, \mathbf{t}_s, \mathbf{X}_p} J^{\prime} =  J + \sum_{s} \sum_{c} \sum_{p \in \mathcal{V}(s,c)} \rho \left( \lambda_s E_b(s,c,p) \right) \\
E_b(s,c,p) = \phi_l \left( [\mathbf{R}_s | \mathbf{t}_s ] [\mathbf{R}_c | \mathbf{t}_c ]^{-1c} \bar{\mathbf{X}}_p, \mathbf{b}_s \right)^2
\end{gathered}
$} ,
\end{equation}
where  $\phi_0$ ($l = 0$) is the loss function for a plane object $\phi_0(\mathbf{X}, \mathbf{b}_s)= \| \max \left( x_3-t_s, 0, -x_3-t_s \right) \|$, and $\phi_1$ ($l = 1$) is the loss function for a cylinder object $\phi_1(\mathbf{X}, \mathbf{b}_s) = \| \sqrt{x_2^2 + x_3^2} - r_s \|$, with an input 3D point $\mathbf{X} = [x_1, x_2, x_3]^{\top}$. The geometric meaning is that after transformation to the object frame, we penalize a 3D point belonging to a cylinder if it is far away from the cylinder surface. Similarly, a 3D point belonging to a plane is penalized if it is out of the boundary of the plane. The weight $\lambda_s$ balances between the cost $J$ of feature points and the cost of semantic object points. In theory, we treat equally both a 3D feature point and an object. As the rotation is defined by its angle-axis, semantic BA is performed by using the LM method with automatic differentiation in Ceres Solver~\cite{agarwal2012ceres}.

\subsection{Measuring Tree Morphology}
In our framework, the trunk diameter estimation is first performed as an input for merging two-sides reconstruction. Canopy-volume and tree-height measurements are conducted based on the merged 3D model of fruit trees, which are illustrated using another dataset captured by stereo infrared cameras from a good view of tree canopies.

\begin{figure}[t]
	\centering
	\includegraphics[width=0.99\columnwidth]{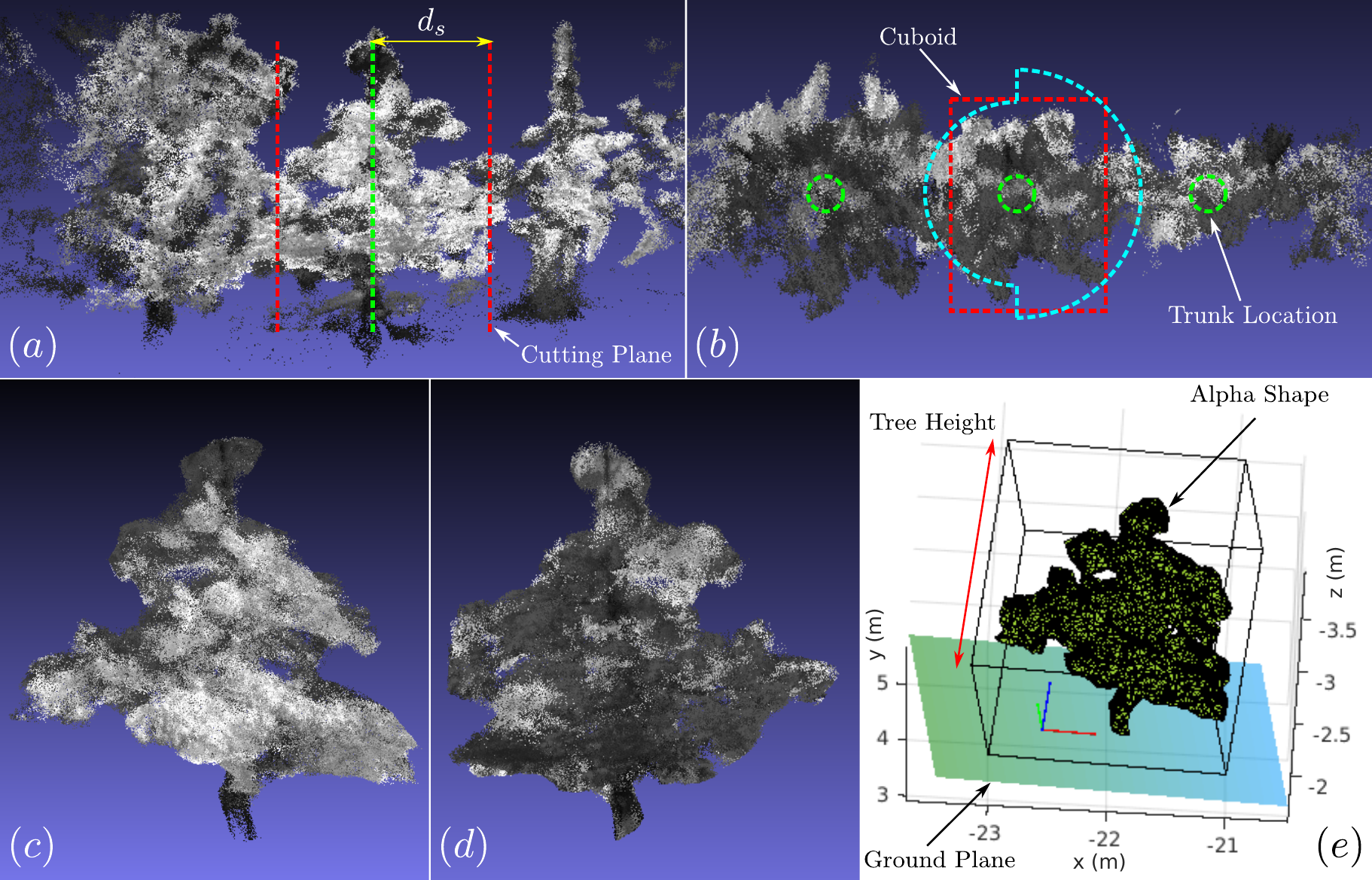}
	\caption{The scheme of estimating canopy volume and tree height. (a): Merged 3D model of a tree row (white front-side points and black back-side points) is partitioned by cutting planes. (b): Top-view tree segmentation based on the union of the cuboid and two-half cylinders. (c) and (d): Segmented tree viewed from both sides. (e): Generated alpha shape with a bounding box on the local ground.}
	\label{fig:treeMorphology}
	\vspace*{-2mm}
\end{figure}

\subsubsection{Trunk Diameter} \label{subsubsec:diameter}
3D dense models of a tree from both sides $\mathcal{F}$ and $\mathcal{B}$ are obtained using volumetric fusion of depth maps from all nearby frames (see Fig.~\ref{fig:fittingOverview}). We first estimate the ground plane as discussed in Sec.~\ref{subsubsec:plane}. The 3D points of the trunk slice are extracted from 3D meshes based on the height to the ground that is determined from annotated 3D points. The trunk diameter is thus robustly estimated from both sides by minimizing the cost
\begin{equation} \label{objectCylinderFB}
\scalebox{0.98}
{$
\begin{gathered}
\argmin_{^{\mathcal{F}}\mathbf{n}_d, ^{\mathcal{B}}\mathbf{n}_d, ^{\mathcal{F}}\mathbf{O}_d, ^{\mathcal{B}}\mathbf{O}_d, r_d} \sum_{p \in \{\mathcal{F},\mathcal{B}\}} e_d^2(\mathbf{X}_p, d) + \lambda \sum_c E_l(c, d) \\
E_l(c, d) =  \| ^c_d\hat{\mathbf{l}}_{\alpha} - ^c\hat{\mathbf{l}}_a \|^2 + \| ^c_d\hat{\mathbf{l}}_{\beta} - ^c\hat{\mathbf{l}}_b \|^2
\end{gathered}
$} ,
\end{equation}
where $^c_d\hat{\mathbf{l}}_{\alpha}$ and $^c_d\hat{\mathbf{l}}_{\beta}$ are two boundary normals of the trunk $d$ in $c$-th annotated frame. The trunk diameter is eventually $2r_d$ which serves as an input in Sec.~\ref{subsubsec:SBA}.

\subsubsection{Canopy Volume}
With a good view of canopies of fruit trees, two-sides 3D reconstructions are first merged in Fig.~\ref{fig:showSBA}. Local grounds are removed given refined semantic information $[\mathbf{R}_s | \mathbf{t}_s ]$. Trunks information indicates the track of the tree row. Based on 3D points distribution~\cite{bargoti2015pipeline}, initial tree segmentation is performed by cutting planes perpendicular to the row track. The cuboid bounding box of a tree is created. From a top view, we assume that a tree is centered at its trunk location projected onto the local ground. To take care of the canopy overlap, the half side of a tree is enclosed by a cylinder with the radius $R_s = \sqrt{2}d_s$, where $d_s$ is the distance from the trunk to the cutting plane (see Fig.~\ref{fig:treeMorphology}). Each tree is thus segmented by taking the union of the bounding box and two-half cylinders. We build an alpha shape~\cite{edelsbrunner1983shape} enclosing all 3D points of each segmented tree by removing small isolated components. The canopy volume is automatically calculated by the alpha-shape algorithm~\cite{edelsbrunner1994three}.

\subsubsection{Tree Height}
Semantic BA outputs optimized information of trunks and local grounds. Based on the trunk location, the pole in the middle of a tree is first segmented out for modern orchards. A bounding box for each tree is then created  to enclose its alpha shape from the local ground plane to the top (see Fig.~\ref{fig:treeMorphology}e). The tree height is thus obtained as the height of the bounding box.

\section{Experiments} \label{sec:experiments}
In this section, we conduct real experiments to evaluate our proposed system for merging 3D mapping of fruit trees from both sides and estimating their morphological parameters.

\begin{figure*}[t]
	\centering
	\includegraphics[width=0.96\textwidth]{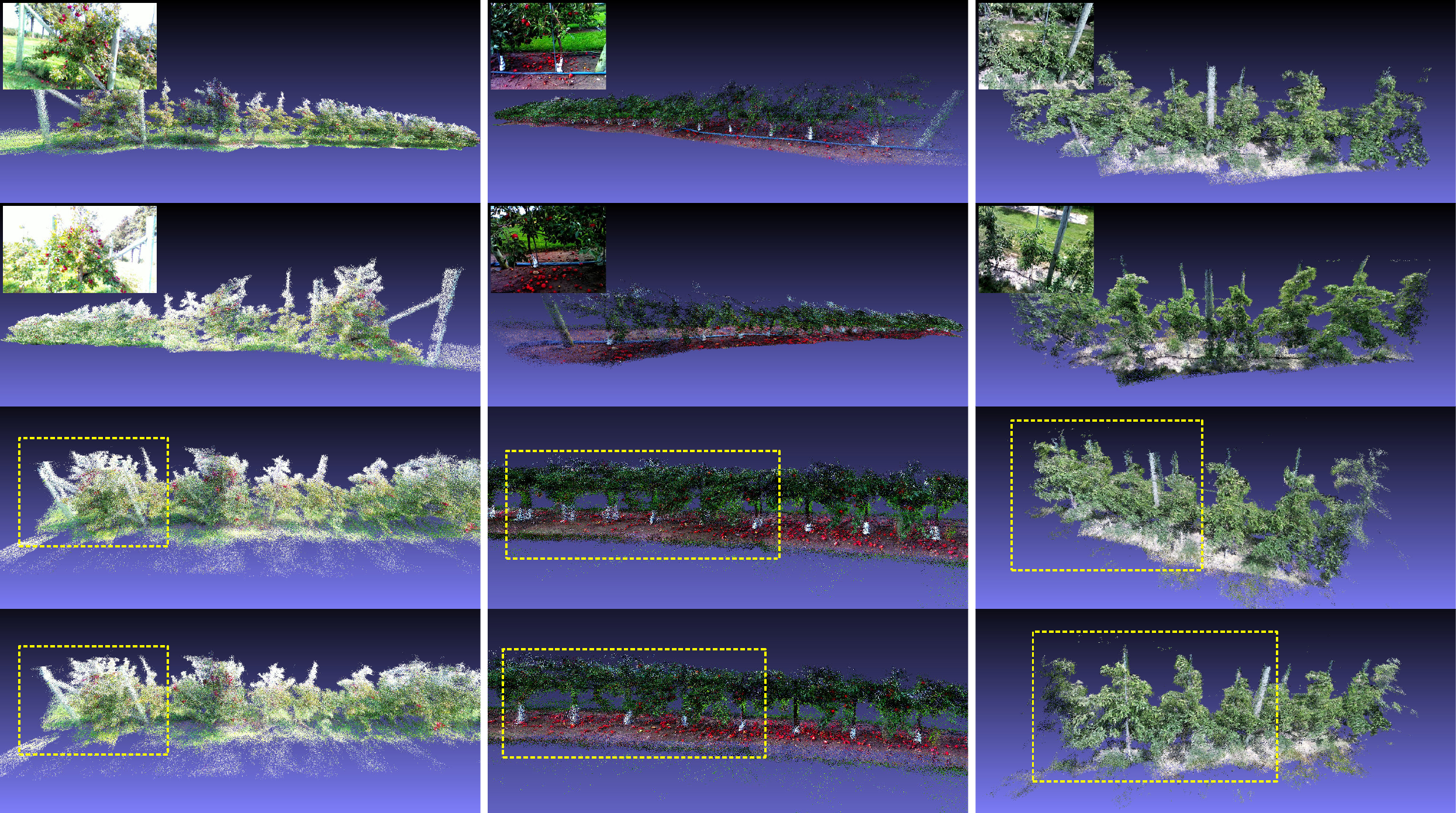}
	\caption{Merging results of 3D reconstruction from both sides of tree rows for Dataset-I, Dataset-II and Dataset-III. Rows 1 and 2: Front-side and back-side 3D reconstructions with scene images. Row 3: Misalignments (yellow boxes) of some landmarks after initial transformation. Row 4: Good 3D models are obtained by eliminating misalignments from semantic BA.}
	\label{fig:experimentMerge}
	\vspace*{-2mm}
\end{figure*}

\begin{figure}[t]
	\centering
	\includegraphics[width=0.96\columnwidth]{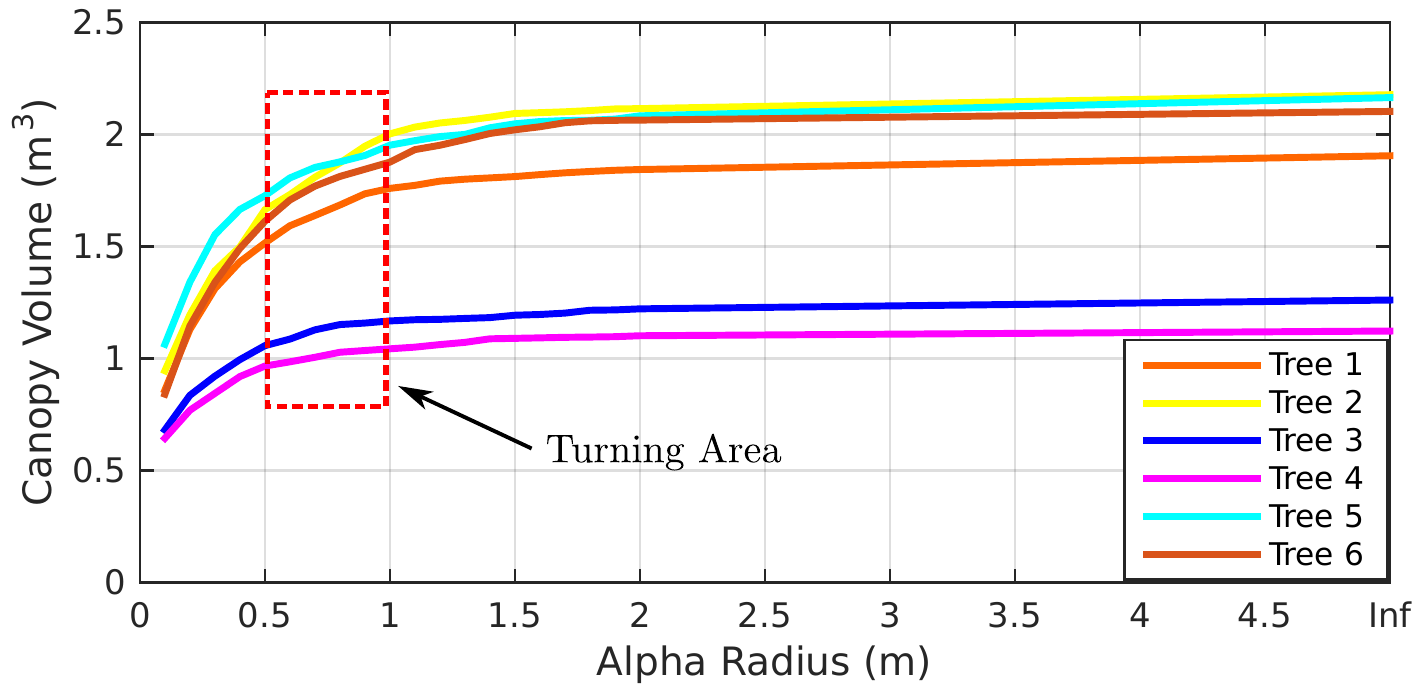}
	\caption{Canopy volumes estimated by alpha shape versus the alpha radius.}
	\label{fig:experimentRadius}
	\vspace*{-2mm}
\end{figure}

\subsection{Datasets and Evaluation Metrics}
The proposed system is tested using three datasets which are all RGB-D data of apple-tree rows in different orchards separately captured from two sides (see Fig.~\ref{fig:experimentMerge}).
Dataset-I is about an apple-tree row with a lot of wild weed captured in a horizontal view. Dataset-II is captured in a tilted view with a focus on tree trunks. Dataset-III is collected by a camera attached to a stick in a tilted-top view of tree canopies. Our merging algorithm is first performed on each dataset, followed by trunk diameter estimation in Dataset-II, and the estimation of canopy volume and tree height in Dataset-III.

To validate the proposed merging algorithm, we first visually check if the misalignment of landmarks (e.g. poles and tree trunks) is eliminated. The objective is to maintain a globally reasonable model of tree rows (from both sides) and while also obtain tree morphology from this 3D information. The accuracy of estimation algorithms are further tested by comparison with manual measurements of trunk diameter and tree height.

\subsection{Implementation Details}
Dataset-I contains 21 trees. Due to the interference of wild weed, only three trunks and three local grounds are used as semantic information for merging algorithm. For Dataset-II, 27 trunks are all annotated with totally 3$\sim$4 frames per each from two sides in order to estimate trunks diameter.
In Dataset-III, a sub-sample of six trees from 30 are chosen for merging demonstration. Since the focus of this dataset is estimating canopy volume and tree height, only three trunks and their local grounds (the middle and two ends) are marked for merging. We use a caliper to measure the actual trunks diameter as the Ground Truth (GT). The GT of trees height and their canopies diameter is obtained by using a measuring stick and a tape, respectively.
\begin{table}[t]
	\centering
	\begin{tabu} to 0.95\columnwidth {| X[2.4,m,c] | | X[m,c] | X[m,c] | X[m,c] | X[m,c] | X[m,c] | X[m,c] |} 
 	\hline
 	\multirow{2}{*}{Model} & \multicolumn{6}{c|}{Section ID of Mean Canopy Volume (m$^3$)}\\
 	\cline{2-7}
 	& V-1 & V-2 & V-3 & V-4 & V-5 & V-6\\
 	\hline\hline
 	Cylinder & 2.957 & 3.105 & 2.503 & 2.185 & 3.155 & 3.307\\
 	\hline
 	Alpha Shape & 1.585 & 1.873 & 1.351 & 1.227 & 1.777 & 1.912\\
 	\hline
 	Convex Hull & 1.805 & 2.177 & 1.460 & 1.322 & 2.064 & 2.202\\
 	\hline
	\end{tabu}
	\caption{Mean canopy volume of 6 tree sections using different models.}
	\label{table:volume}
 	\vspace*{-6mm}
\end{table}

\subsection{Morphology Estimation Results}
\subsubsection{Merging 3D Reconstruction}
As shown in Fig.~\ref{fig:experimentMerge}, the proposed method is able to build a well-aligned global 3D models of tree rows even without annotation for each tree. Specifically, duplicated poles and trunks are all merged. In general, the merging algorithm only requires two-sides object correspondences around two ends and the middle of each tree row. When there is no need for estimating trunks diameter, we can roughly annotate a long section of a trunk as a cylinder, or even other landmarks, such as supporting poles and stakes. The planar assumption of local ground for each tree makes general our method which can be applied to any orchard environments without concern about the terrain.

\subsubsection{Comparison and Analysis}
In Dataset-II, we select 14 trees among 27 to demonstrate in detail the accuracy of our algorithm for trunks diameter estimation. If without 2D constraints, trunk diameters are always estimated larger than GT due to unreliable depth values around scene boundaries. Table~\ref{table:diameter} shows that 
with 2D constraints the average error of our diameter estimation is around 5 mm. For small trunks, the estimated results are still larger than GT, since the camera is relatively far from small trunks. Large pixel errors of edge detection (low resolution for trunk boundaries) thus cause the diameter overfitting. It implies that the camera should closely capture these trees with small trunks. In Dataset-III, we perform tree height estimation for 14 trees chosen among 30. Table~\ref{table:height} shows that the average error of our tree height estimation is around 4 cm. The estimation results for trunk diameter and tree height thus demonstrate the high accuracy of the proposed vision system.
\begin{figure*}[t]
	\centering
	\includegraphics[width=0.96\textwidth]{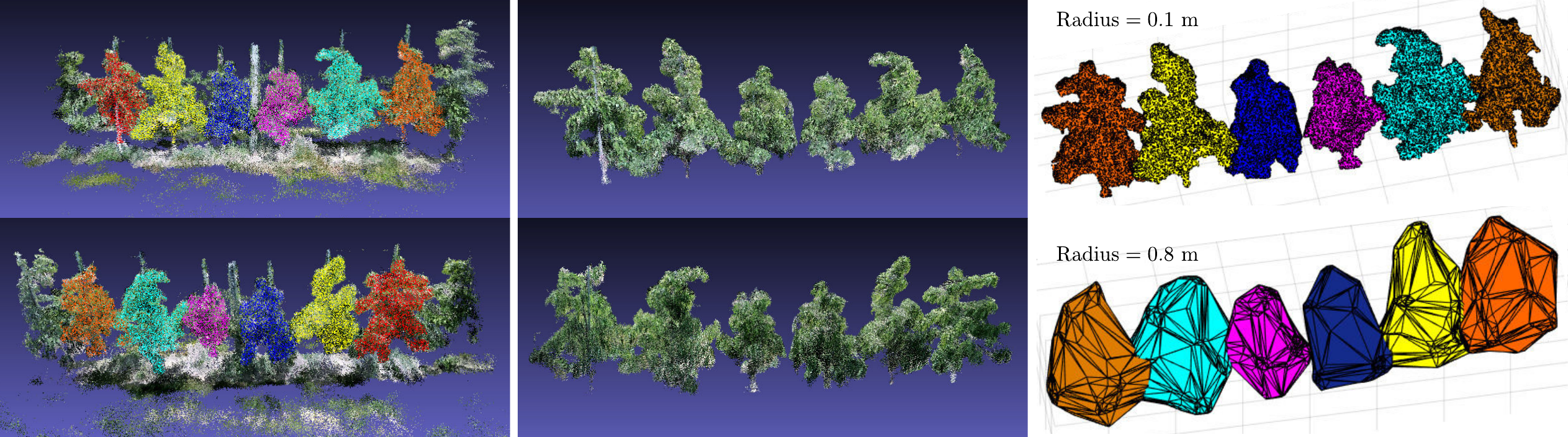}
	\caption{Six sample trees in Dataset-III are segmented and enclosed by alpha shapes. Column 1: Each tree is differentiated from front-side and back-side reconstructions. Column 2: Six trees are segmented out from both sides. Colume 3: Alpha shapes of six trees are generated using two different alpha radiuses from two-side views.}
	\label{fig:experimentShape}
	\vspace*{0mm}
\end{figure*}
\begin{table*}[t]
	\centering
	\begin{tabu} to 0.99\textwidth {| X[2.1,c,m] | | X[c,m] | X[c,m] | X[c,m] | X[c,m] | X[c,m] | X[c,m] |  X[c,m] | X[c,m] | X[c,m] | X[c,m] | X[c,m] | X[c,m] | X[c,m] | X[c,m] | | X[c,m] |} 
 	\hline
 	Tree ID & T-2 & T-4 & T-6 & T-8 & T-9 & T-11 & T-13 & T-15 & T-18 & T-19 & T-22 & T-24 & T-26 & T-27 & Mean\\
 	\hline\hline
 	Est. & 5.24 & 5.10 & 5.48 & 8.04 & 6.56 & 6.50 & 5.51 & 5.87 & 5.29 & 5.70 & 5.99 & 5.49 & 5.77 & 5.37 & $-$\\
 	\hline
 	GT & 5.39 & 4.12 & 4.77 & 8.22 & 6.68 & 6.82 & 5.08 & 5.23 & 4.37 & 5.00 & 5.70 & 5.63 & 5.24 & 4.61 & $-$\\
 	\hline
 	Error (cm) & 0.15 & 0.98 & 0.74 & 0.18 & 0.12 & 0.32 & 0.43 & 0.64 & 0.92 & 0.70 & 0.29 & 0.14 & 0.53 & 0.76 & 0.49\\
 	\hline
	\end{tabu}
	\caption{Estimation errors of trunk diameter in Dataset-II.}
	\label{table:diameter}
 	\vspace*{-4mm}
\end{table*}
\begin{table*}[t]
	\centering
	\begin{tabu} to 0.99\textwidth {| X[2.1,c,m] | | X[c,m] | X[c,m] | X[c,m] | X[c,m] | X[c,m] | X[c,m] |  X[c,m] | X[c,m] | X[c,m] | X[c,m] | X[c,m] | X[c,m] | X[c,m] | X[c,m] | | X[c,m] |}
 	\hline
 	Tree ID & H-1 & H-2 & H-3 & H-4 & H-5 & H-6 & H-7 & H-16 & H-18 & H-19 & H-20 & H-21 & H-22 & H-23 & Mean\\
 	\hline\hline
 	Est. & 2.145 & 2.050 & 2.453 & 2.463 & 2.131 & 1.997 & 2.087 & 2.357 & 2.456 & 2.311 & 1.990 & 2.084 & 2.496 & 2.361 & $-$\\
 	\hline
 	GT & 2.159 & 2.032 & 2.362 & 2.515 & 2.083 & 1.981 & 2.108 & 2.438 & 2.413 & 2.337 & 2.032 & 2.057 & 2.489 & 2.413 & $-$\\
 	\hline
 	Error (m) & 0.014 & 0.018 & 0.091 & 0.052 & 0.048 & 0.016 & 0.021 & 0.081 & 0.043 & 0.026 & 0.042 & 0.027 & 0.007 & 0.052 & 0.038\\
 	\hline
	\end{tabu}
	\caption{Estimation errors of tree height in Dataset-III.}
	\label{table:height}
 	\vspace*{-8mm}
\end{table*}

In Dataset-III, we first segment out six sample trees and generate enclosing alpha shapes (see Fig.~\ref{fig:experimentShape}) to represent their canopies. However, the alpha radius should be appropriately chosen. The alpha shape with a small radius value will produce holes inside the canopy, which is not desirable form the view of horticultural study. Fig.~\ref{fig:experimentRadius} shows that the canopy volume increases and converges to a constant value as the alpha radius increases to infinity, which produces a convex hull. The best value of alpha radius should represent a canopy model without holes and produce the smallest volume. Thus, we set the radius as $0.8$ m within the turning area (See Fig.~\ref{fig:experimentRadius} and Fig.~\ref{fig:experimentShape}).

One of the common methods used in horticultural science for modeling canopies is to treat a tree as a cylinder. To show the difference among different models of canopies, we divide 18 trees from Dataset-III into 6 sections based on their relatively similar sizes, and report the mean canopy volume of each section in Table~\ref{table:volume}. It should be noticed that simple cylinder model overestimates the canopy volume. Thus, it is reasonable to consider that our proposed method for canopy volume estimation is more suitable to generalize the geometry of tree structures, which is promising to build the ground truth of tree canopies for horticulturists using the proposed vision system.


\section{Conclusion and Future Work} \label{sec:conclusion}
In this work, we presented a vision system that collects RGB and depth images of fruit trees in the orchard, and uses this information to estimate morphological parameters for phenotyping, such as tree volume, tree height and trunk diameter. Our system consists of an RGB-D camera attached to a stick, which can be further mounted on a moving platform. 3D models of fruit trees from both sides are generated separately and merged into a global model by exploiting semantic information (i.e. trunk region of interest and local ground). Tree volume can be immediately computed based on partitioned model of each tree refined by our algorithm. We also proposed robust fitting algorithms for estimating tree height and trunk diameter. Our system is  evaluated using three different types of tree datasets collected in orchards. This is the first vision system that can measure morphological parameters of trees in fruit orchards by using only an RGB-D camera. Future work will focus on automated extraction of semantic information, such as trunk detection and tree separation in densely packed scenario. Moreover, merged model of fruit trees from both sides will be used for fruit tracking in 3D to avoid double counting.

The only assumption in the proposed method is that we are given the data association of object correspondences from two sides (i.e. correct matching tree indices from both sides of a tree row). 
For the reason of high accuracy, we annotate the trunk silhouette for measuring its diameter. If without the need for accurate diameter estimation, manual annotation can be replaced by automatic object detection~\cite{salas2013slam++}. The data-association assumption can be further removed by developing a stable technique to detect and segment out individual plant in agricultural environments. We will be working on these improvements in our future work.


\bibliographystyle{plainnat}
\bibliography{references}

\end{document}